%% file: main_arxiv.tex
\documentclass[10pt,twocolumn,letterpaper]{article}

\usepackage[pagenumbers]{cvpr} 

\input{preamble}

\definecolor{cvprblue}{rgb}{0.21,0.49,0.74}
\usepackage[pagebackref,breaklinks,colorlinks,citecolor=cvprblue]{hyperref}

\usepackage{color}
\usepackage{colortbl}
\usepackage{xcolor}
\usepackage[accsupp]{axessibility}
\usepackage{array}
\usepackage{graphicx}
\usepackage{amsmath}
\usepackage{amssymb}
\usepackage{booktabs}
\usepackage{makecell}
\usepackage{framed,multirow}
\usepackage{threeparttable}
\usepackage{tabulary}
\usepackage{multirow,multicol}
\usepackage{soul}

\definecolor{DnCBG}{rgb}{0.9, 0.9, 1.}
\definecolor{Gray}{gray}{0.5}
\definecolor{GrayBG}{gray}{0.92}
\newlength\savewidth\newcommand\shline{\noalign{\global\savewidth\arrayrulewidth
  \global\arrayrulewidth 0.8pt}\hline\noalign{\global\arrayrulewidth\savewidth}}

\newcommand{\blue}[1]{\textcolor[RGB]{61,90,128}{#1}}

\def\paperID{3751} 
\def\confName{CVPR}
\def\confYear{2024}

\title{Representing Part-Whole Hierarchies in Foundation Models by Learning Localizability, Composability, and Decomposability \\ from Anatomy via Self-Supervision}

\author{Mohammad Reza Hosseinzadeh Taher$^{1}$ \quad
Michael B. Gotway$^2$ \quad Jianming Liang$^1$\\
$^1$Arizona State University \quad $^2$Mayo Clinic\\
{\tt\small \{mhossei2,jianming.liang\}@asu.edu} \quad {\tt\small Gotway.Michael@mayo.edu} 
}
\begin{document}

\twocolumn[{
\renewcommand\twocolumn[1][]{#1}
\maketitle
\vspace{-230pt}

\begin{center}
    \centering
    \textcolor{red}{Please cite the paper as}\\ \textcolor{blue}{Mohammad Reza Hosseinzadeh Taher, Michael B. Gotway, and Jianming Liang. Representing Part-Whole Hierarchies in Foundation Models by Learning Localizability, Composability, and Decomposability from Anatomy via Self-Supervision. In Proceedings of the IEEE/CVF Conference on Computer Vision and Pattern Recognition (CVPR), 2024.}
\end{center}
\vspace{164pt}
}]

\input{sec/0_abstract}    
\input{sec/1_intro}

\input{sec/2_method}

\input{sec/3_Implementation}

\input{sec/4_results}
\input{sec/5_related_work}

\input{sec/6_conclusion}
\input{sec/7_appendix}
{
    \small

}

\end{document}

%% file: preamble.tex
\usepackage[dvipsnames]{xcolor}


%% file: sec/0_abstract.tex
\begin{abstract}
\noindent Humans effortlessly interpret images by parsing them into part-whole hierarchies; deep learning excels in learning multi-level feature spaces, but they often lack explicit coding of part-whole relations, a prominent property of medical imaging. To overcome this limitation, we introduce Adam--v2, a new self-supervised learning framework extending Adam~\cite{Taher2023Adam} by explicitly incorporating part-whole hierarchies into its learning objectives through three key branches: (1) Localizability, acquiring discriminative representations to distinguish different anatomical patterns; (2) Composability, learning each anatomical structure in a parts-to-whole manner; and (3) Decomposability, comprehending each anatomical structure in a whole-to-parts manner. Experimental results across 10 tasks, compared to 11 baselines in zero-shot, few-shot transfer, and full fine-tuning settings, showcase Adam--v2's superior performance over large-scale medical models and existing SSL methods across diverse downstream tasks. The higher generality and robustness of Adam--v2’s representations originate from its explicit construction of hierarchies for distinct anatomical structures from unlabeled medical images. Adam--v2 preserves a semantic balance of anatomical diversity and harmony in its embedding, yielding representations that are both generic and semantically meaningful,  yet overlooked in existing SSL methods. All code and pretrained models are available at  \href{https://github.com/JLiangLab/Eden}{GitHub.com/JLiangLab/Eden}.

\end{abstract}

%% file: sec/1_intro.tex
\section{Introduction}
\label{sec:intro}

\begin{figure} [t]
    \centering
    \includegraphics[width=0.95\linewidth]{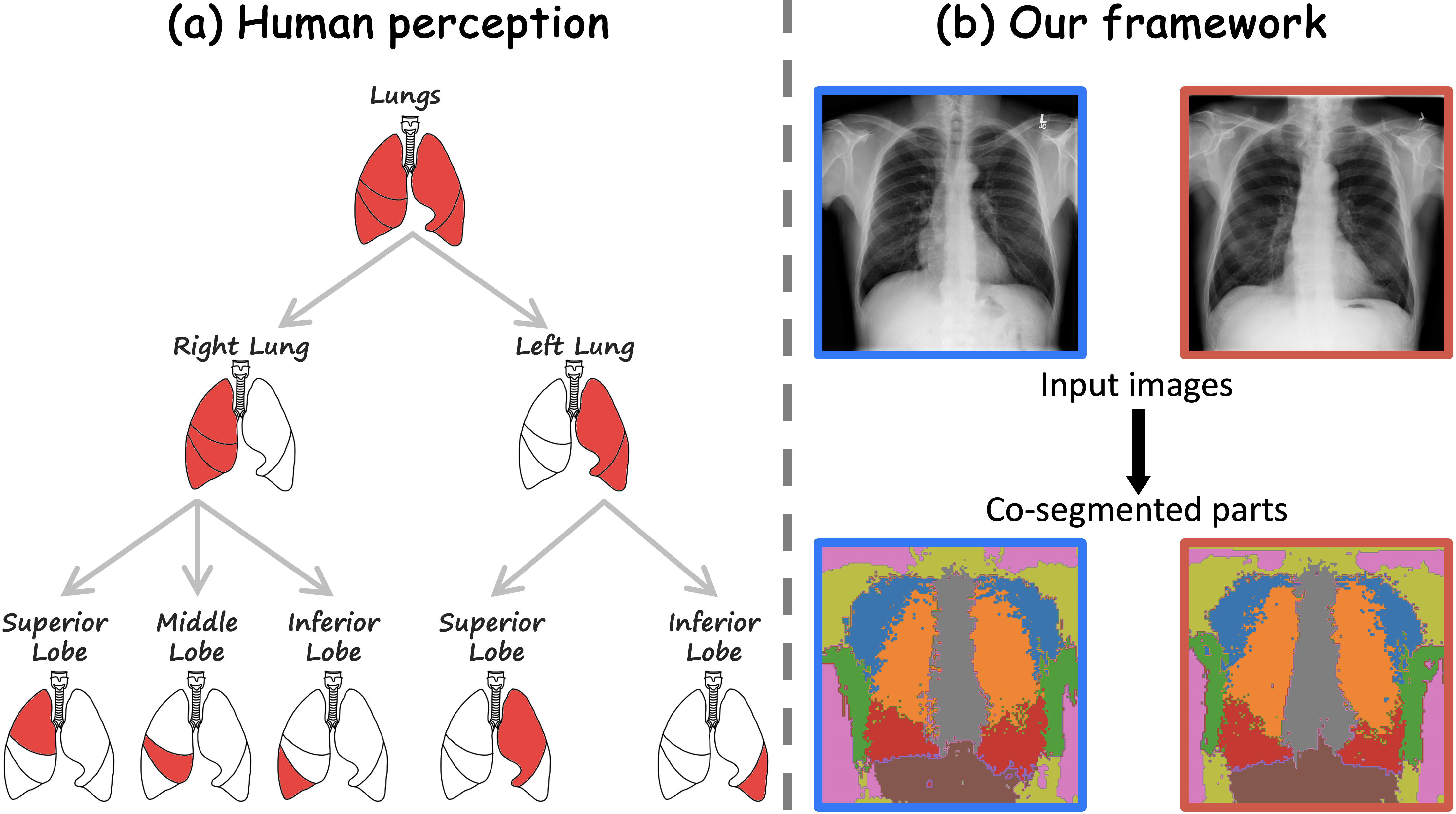}
\caption{Human perception effortlessly organizes objects into hierarchies to understand their part-whole relationships in images. Taking lungs as an example in (a), even a non-radiologist can form a hierarchy of the right and left lungs, whereas a radiologist can further ``see’’ the lobes in sub-hierarchies. To emulate this ability, we introduce a self-supervised learning framework that \textit{explicitly} learns to encode inherent part-whole hierarchies within medical images into an embedding space, leading to the development of a powerful model (Adam--v2) that is foundational to medical imaging. Adam-v2 can transform each pixel in medical images (e.g., chest radiographs in (b)) into semantically meaningful embeddings (Eve--v2), forming multiple ``\textit{echo chambers}’’ (produced via co-segmentation~\cite{amir2021deep,zhou2023learning})---different anatomical structures are associated with distinct embeddings, and the same anatomical structures have (nearly) identical embeddings across patients.}
    \label{fig:intuition}
\end{figure}

Human perception effortlessly parses visual scenes into \textit{part-whole} hierarchies~\cite{Hinton1979SomeDO,HINTON1990Mapping,hinton2021represent}. For instance, when interpreting a chest radiograph, even untrained observers can quickly form a hierarchy by dividing the lower respiratory tract into the left and right lungs, whereas more experienced observers can invoke further sub-hierarchies (see \cref{fig:intuition}). Deep learning has enabled breakthroughs in learning visual representation at multiple levels. However, the multi-level feature space learned by deep models does not explicitly code part-whole hierarchies with necessary semantic information to indicate hierarchical relationships among wholes and their constituent parts~\cite{mounir2023streamer,hinton2021represent}.

To mimic the human ability to understand part-whole hierarchies in images, Hinton, in his \ul{\textit{idea}} paper~\cite{hinton2021represent}, introduced an \ul{\textit{imaginary}} system (\textit{i.e.}, GLOM), aiming to signify the importance of explicitly presenting part-whole hierarchies in a neural network. Inspired by the conceptual idea underlying GLOM, we devise a novel self-supervised learning (SSL) framework, leading to a \ul{\textit{functioning}} system that, from medical images, \textit{autodidactically} constructs a hierarchy of embeddings for distinct anatomical structures, semantically balancing anatomical diversity and harmony at each level and conveying parental ``whole'' at the higher level and filial ``parts'' at the lower level.

Our framework, as illustrated in~\cref{fig:method}, comprises three branches: (1) ``localizability'', which compels the model to learn a semantically structured embedding space by discriminating between different anatomical structures, (2) ``composability'', which empowers the model to learn part-whole relations by constructing each anatomical structure through the integration of its constituent parts, and (3) ``decomposability'', which encourages the model to learn whole-part relations by decomposing each anatomical structure into its constituent parts. Unifying these three branches together in a coarse to fine learning approach, the localizability branch enables the model to preserve harmony in embeddings of semantically similar anatomical structures in a hierarchy of scales. Simultaneously, composability and decomposability branches empower the model to not only convey hierarchical relationships but also preserve diversity of semantically similar anatomical structures across patients through encoding finer-grained anatomical information of their constituent parts. We call our system (\textit{i.e.}, pretrained model) \textbf{Adam--v2} because it represents a significant advancement from our previous version---Adam (\textbf{a}utodidactic \textbf{d}ense \textbf{a}natomical \textbf{m}odels)~\cite{Taher2023Adam}---that learns \textit{autodidactically} and yields dense anatomical embedding, nicknamed \textbf{Eve--v2} (\textbf{e}mbedding \textbf{ve}ctors) for semantic richness. We further coin our project site \textbf{Eden} (\textbf{e}nvironment for \textbf{d}ense \textbf{e}mbeddings and \textbf{n}etworks), where all code, pretrained Adam--v2, and Eve--v2 are placed.

We extensively evaluate Adam--v2 in \textbf{(1)} \textit{Zero-shot setting} (\S\ref{sec:results_anatomy_understanding}): Adam--v2 yields more semantically meaningful embeddings (Eve--v2) compared with existing SSL methods with a set of unique properties essential for anatomy understanding (\cref{fig:locality,fig:hierachy,fig:composition,fig:extra_intra}); \textbf{(2)} \textit{Few-shot transfer setting} (\S\ref{sec:results_fewshot}): Adam--v2 outperforms 2 \textit{large-scale} medical models, RadImageNet and LVM-Med, as well as a representative set of  7  SSL methods by a remarkable margin in anatomical structure and disease segmentation tasks (\cref{tab:fewshot}); and \textbf{(3)} \textit{Full fine-tuning setting} (\S\ref{sec:results_fullfinetune}): Adam--v2 provides more generalizable representations compared to fully-supervised and SSL baselines across a myriad of tasks (\cref{fig:sota_full_finetune} \& \cref{tab:public_chestxray14}). Our main contributions are as follows:

\begin{itemize}
    \item A new self-supervised learning strategy, called Adam--v2, that encodes inherent hierarchical relationships within medical images, yielding discriminative representations blended with semantics of part-whole relations.
    
    \item A comprehensive set of experiments proves higher generalizability and robustness of Adam--v2, particularly highlighting Adam--v2’s proficiency in few-shot transfer and achieving a new record in the ChestX-ray14 benchmark.
    
    \item A set of quantitative and qualitative feature analyses that opens up novel perspectives for assessing anatomy understanding from various viewpoints.
\end{itemize}

%% file: sec/2_method.tex
\section{Method}
\label{sec:method}

\begin{figure*}
    \centering
    \includegraphics [width=0.88\linewidth]{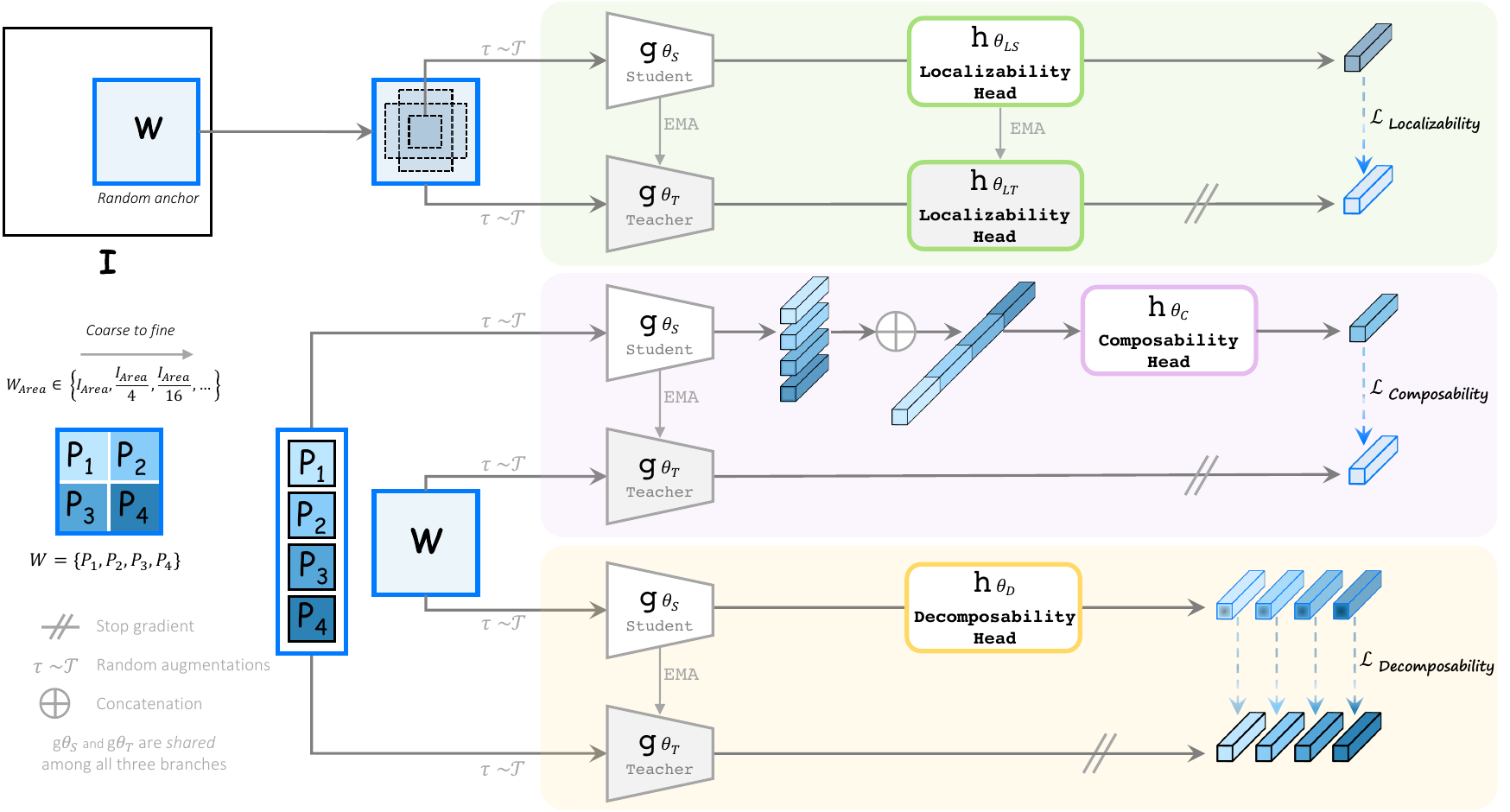}
    \caption{Adam--v2 learns hierarchical representations in a coarse-to-fine-manner via three branches: localizability, composability, and decomposability. Given an anchor whole $w$ randomly sampled from image $I$, the localizability branch augment and process $w$ and its multi-scale views, and enforce consistency between their embeddings, yielding distinct features for different anatomical structures. The composability branch  decomposes $w$ into a set of parts, and enforces consistency between the embedding of $w$ and the aggregated embeddings of its parts, encoding \textit{part-whole} relations. The decomposability branch decomposes the embedding of $w$ to acquire the embeddings of its constituent parts, and enforce consistency between the embeddings of parts and their decomposed counterparts, capturing \textit{whole-part} relations.}
    \label{fig:method}
\end{figure*}

Our framework, depicted in \cref{fig:method}, aims to underpin the development of powerful self-supervised models foundational to medical imaging by constructing a hierarchy of embeddings learned from anatomy. Our framework comprises three key branches: (1) localizability, aiming to acquire discriminative representations for distinguishing different anatomical structures; (2) composability, aiming to learn each anatomical structure in a parts-to-whole manner; and (3) decomposability, aiming to comprehend each anatomical structure in a whole-to-parts manner. Seamlessly integrating these learning objectives into a unified framework captures inherent part-whole hierarchies within medical images, yielding a powerful model (Adam--v2) that can serve not only as the foundation for myriad target tasks via adaptation (fine-tuning), but also its embedding vectors (Eve--v2) bear rich semantics, usable standalone without adaptation (zero-shot), for other tasks like landmark detection. The following details our framework.

\subsection{Learning Localizability}
\label{sec:localizability}
The localizability branch seeks to learn  a semantically-structured embedding space where similar anatomical structures
are clustered together and are distinguished from dissimilar anatomical structures. As illustrated in \cref{fig:method}, the localizability branch includes  the student $g_{\theta_S}$ and teacher $g_{\theta_T}$ encoders, and two projectors  $h_{\theta_{LS}}$ and  $h_{\theta_{LT}}$, referred to as localizability heads. The  parameters  of student $g_{\theta_S}$ and localizability head  $h_{\theta_{LS}}$  are learned  with stochastic gradient descent while the parameters of the teacher $g_{\theta_T}$ and  head $h_{\theta_{LT}}$ are updated using an exponential moving average (EMA) on the weights of  $g_{\theta_S}$ and $h_{\theta_{LS}}$, respectively. Given an anchor patch $w$ randomly sampled from the input image $I$, we extract a set  $C$ of multi-scale crops from $w$. In particular, these crops exhibit diverse dimensions while sharing the same or slightly shifted center as $w$, contributing to a comprehensive understanding of the same anatomical structure at various resolutions. We then apply random data augmentations $\scriptstyle{\mathcal{T}(.)}$ on $w$ and multi-scale crops in $C$. The augmented view of $w$ is passed to the teacher, while the augmented views of the crops in $C$ are passed to the student network, generating the features $y_t=g_{\theta_T}({\scriptstyle\mathcal{T}}(w))$ and  $Y_s=\{g_{\theta_S}({\scriptstyle\mathcal{T}}(c)) \;|\; c\in C \}$, respectively. The localizability heads project the features to the output embeddings  $z_t = h_{\theta_{LT}}(y_t) $ and $Z_s=\{h_{\theta_{LS}}(y_s) \;|\; y_s\in Y_s\}$, which are normalized
with a softmax function:
\begin{equation}
  P_t(z_t)^{(i)} = \frac{\exp(z_t^{(i)} / \tau_t)}{\sum_{k=1}^K \exp(z_t^{(k)} / \tau_t)}
\end{equation}    
where $\tau_t>0$ is a temperature parameter controlling the sharpness of the output distribution, and $K$ is the output dimension of the localizability heads. A  softmax function $P_s$ with temperature $\tau_s$ is similarly employed to normalize the features in $Z_s$. The localizability branch’s objective is to maximize the consistency between the embeddings  of the input anchor and its augmented views. To do so, we employ cross-entropy loss~\cite{Caron2021Emerging}: 
\begin{equation}
	\mathcal{L}_{Localizability} =  - \frac{1}{|Z_s|}\sum_{z_s \in Z_s} P_t(z_t) \log P_s(z_s)
\label{eq:loss}
\end{equation}
It is noteworthy that our framework offers flexibility in utilizing various localizability loss functions. While we opt for a self-distillation loss due to its simplicity and efficiency~\cite{Grill2020Bootstrap,Caron2021Emerging,Song2023Multi}, alternative sophisticated objectives, such as contrastive loss~\cite{He2020Momentum,Chen2020Simple}, can also be employed.

\subsection{Learning Composability}
\label{sec:composability}
The composability branch seeks to learn the \textit{part-whole} anatomical hierarchies in a bottom-up manner by assembling larger anatomical structures from their smaller constituent subparts. As illustrated in \cref{fig:method}, the composability branch consists of the student $g_{\theta_S}$ and teacher $g_{\theta_T}$ encoders, which are shared with the localizability branch, and a composability head $h_{\theta_{C}}$. Given an anchor whole $w$ randomly sampled from the input image $I$, we decompose it into a set of $n$ non-overlapping parts $\mathrm{P}=\{p_i\}_{i=1}^n$. The parts are augmented and processed by the student network, generating parts' embeddings $Y_{ps} = \{y_i = g_{\theta_S}({\scriptstyle\mathcal{T}}(p_i))\}_{i=1}^n$. The parts' embeddings are then concatenated and passed to the composability head $h_{\theta_{C}}$ to produce the aggregated embeddings of parts $z_{ps} = h_{\theta_{C}} ( \oplus(\{y_i\}_{i=1}^n))$. Moreover, the whole anatomical structure $w$ is augmented and passed to the teacher network to generate the whole's embedding $z_{wt} = g_{\theta_T}({\scriptstyle\mathcal{T}}(w))$. The composability branch is trained to maximize the agreement between the whole's embedding and the the aggregated embeddings of its parts: 
\begin{equation}
  \mathcal{L}_{Composability} = \ell_s(z_{wt}, z_{ps}) 
  \label{eq:loss_composability}
\end{equation}
where $\ell_s(z_{wt}, z_{ps})$ presents a function that measures similarity between $z_{wt}$ and $z_{ps}$, such as MSE~\cite{Grill2020Bootstrap}, cross-entropy~\cite{Caron2021Emerging}, or cosine similarity~\cite{Chen2021Exploring}.

\subsection{Learning Decomposability}
\label{sec:decomposability}
The decomposability branch seeks to learn the \textit{whole-part} anatomical hierarchies in a top-down manner by decomposing larger anatomical structures into their smaller constituent subparts. As shown in \cref{fig:method}, the decomposability branch comprises the student $g_{\theta_S}$ and teacher $g_{\theta_T}$ encoders, which are shared with the localizability and composability branches, and a decomposability head $h_{\theta_{D}}$. Given an anchor whole $w$, we decompose it into a set of 
 $n$ non-overlapping parts $\mathrm{P}=\{p_i\}_{i=1}^n$. The anchor whole $w$ is augmented and fed into the student network, producing the whole's embedding $z_{ws} = g_{\theta_S}({\scriptstyle\mathcal{T}}(w))$. The whole's embedding is then passed to the decomposability head $h_{\theta_{D}}$, which decomposes it into a set of individual embeddings corresponding to the constituent parts of the whole $Z_{ps} = h_{\theta_{D}}(z_{ws})$. 
Additionally, the parts $\mathrm{P}=\{p_i\}_{i=1}^n$ are augmented and processed by the teacher network, generating parts' embeddings $Z_{pt} = \{g_{\theta_T}({\scriptstyle\mathcal{T}}(p_i))\}_{i=1}^n$. The decomposability branch is trained to maximize the agreement between the embeddings of the individual parts and their decomposed counterparts:
\begin{equation}
  \mathcal{L}_{Decomposability} = \frac{1}{|P|} \sum_{i=1}^{|P|}  \ell_s(z_{p_i}, z_{p'_i})
  \label{eq:loss_decomposability}
\end{equation}
where $z_{p_i} \in Z_{pt}$ and $z_{p'_i}\in Z_{ps}$, and $\ell_s(z_{p_i}, z_{p'_i})$ presents a function that measures similarity between $z_{p_i}$ and $z_{p'_i}$, such as MSE, cross-entropy, or cosine similarity.

\subsection{Training Pipeline}
To  guide the model in learning hierarchical representations, we consider a hierarchy of diverse anatomical structures at various scales. Specifically, the highest level of the hierarchy represents entire images (of spatial resolution ${\scriptstyle(H\times W)}$) with complete anatomy, while each subsequent level $m$ $\in\{1,2 ...\}$ represents anatomical structures $w$ at a scale of $(\frac{H}{2^m}\times\frac{W}{2^m})$, randomly sampled from the images. In a coarse to fine manner, the anatomical structures $w$ at each level are fed as the input to the localizability, composability, and decomposability branches, and are learned through the following combined loss function:
\begin{equation}\label{eq:loss}
\resizebox{0.9\hsize}{!}{$
  \mathcal{L} = \lambda_{1}*\mathcal{L}_{Localizability} + \lambda_{2} * \mathcal{L}_{Composability} + \lambda_{3} * \mathcal{L}_{Decomposability}$ }
\end{equation}
where $\lambda_1$, $\lambda_2$, $\lambda_3$ are coefficients denoting the weight of each loss term. Through our unified training scheme, Adam--v2 learns a rich embedding space that preserves harmony among similar anatomical structures and encoding their hierarchical relations.

%% file: sec/3_Implementation.tex
\section{Implementation Details}
\label{sec:implementation}

\noindent\textbf{Pretraining protocol.} We use \textit{unlabeled} chest radiographs and color fundus photographs for pretraining Adam--v2 on two imaging modalities. Our SSL framework is architecture-neutral and compatible with any ConvNet and vision transformer backbones. As an illustration, we pretrain Adam--v2 with ResNet-50~\cite{he2016deep}, ViT-S~\cite{dosovitskiy2020vit}, and ConvNeXt-B~\cite{liu2022convnet} backbones. We follow~\cite{Caron2021Emerging} in optimization settings (e.g. optimizer, learning rate schedule, $\tau_t$, $\tau_s$, etc), updating teacher weights, and architecture of $h_{\theta_{LS}}$ and  $h_{\theta_{LT}}$ heads.  $h_{\theta_{C}}$ and $h_{\theta_{D}}$ are two-layer MLP heads. We use MSE as $\ell_s(.)$ in \cref{eq:loss_composability,eq:loss_decomposability}. $\lambda_1$, $\lambda_2$, $\lambda_3$ are set to  1, $n$ to 4, and  $m$ up to 4. In localizability branch, following~\cite{Caron2021Emerging,caron2021unsupervised}, we extract one $224^2$ global view and eight $96^2$ multi-scale crops from $w$. For other branches, we use input resolution $224^2$. Augmentation ${\scriptstyle\mathcal{T}(.)}$ includes color jittering, Gaussian blur, and rotation. To prove the scalability of our framework, we train a large-scale model using ConvNeXt-B backbone and a large corpus of  926,028 images collected from 13 different public chest X-ray datasets. 

\smallskip
\noindent\textbf{Evaluations.} We evaluate our framework in zero-shot, few-shot, and full transfer settings. We consider 10 downstream tasks on 9 publicly available datasets for fine-tuning settings, including 
JSRT~\cite{vanginneken2006Segmentation}, VinDR-Rib~\cite{Nguyen2021VinDr}, ChestX-Det~\cite{Lian2021Structure}, SIIM-ACR~\cite{PNEchallenge},  VinDr-CXR~\cite{nguyen2020vindrcxr}, NIH Shenzhen~\cite{Jaeger2014Tow}, ChestX-ray14~\cite{wang2017chestx},  DRIVE~\cite{Budai2013Robust}, and Drishti-GS~\cite{Sivaswamy2014Drishti}. These tasks rigorously evaluate the generalizability of our Adam--v2 across a range of applications, diseases, anatomical structures, and modalities. 

\smallskip
\noindent\textbf{Baselines.} 
We compare Adam--v2 with a representative set of seven SOTA publicly-available SSL baselines, encompassing ConvNet- and transformer-based methods. These baselines represent diverse objectives at instance-, patch-, and pixel-level, among which 
TransVW~\cite{haghighi2021transferable},  PCRL~\cite{Zhou2021Preservational}, DiRA~\cite{Haghighi2022DiRA}, and Medical-MAE~\cite{Xiao2023Delving} represent SOTA methods tailored for medical tasks. All SSL baselines are pretrained on the same datasets as our Adam--v2 by following their official settings. Moreover, we compare Adam--v2 with the publicly available and official models of two recent \textit{large-scale} medical models: RadImageNet~\cite{Mei2023RadImageNet} and LVM-Med~\cite{nguyen2023lvmmed}, pretrained on 1.3 million medical images in fully-supervised and self-supervised manners, respectively.

\smallskip
\noindent\textbf{Fine-tuning protocol.} 
Following the standard transfer learning protocol~\cite{Taher2021Systematic}, Adam--v2’s pretrained teacher model has been fine-tuned for (1) classification tasks by appending a task-specific head, and (2) segmentation tasks by utilizing a U-Net network~\cite{Ronneberger2015Unet}, where the encoder is initialized with the pretrained weights. We run each method for each task at least five times. We provide statistical analysis using an independent two-sample $t$-test.

%% file: sec/4_results.tex
\section{Results and Analysis}
\label{sec:results}

\subsection{Adam--v2 demonstrates zero-shot anatomy understanding, offering semantics-rich embeddings over existing SSL methods}
\label{sec:results_anatomy_understanding}

This section showcases the anatomy understanding capabilities of our framework by delving into the unique \textit{learned} and \textit{emergent} properties of our Adam--v2's embeddings (Eve--v2) in various \ul{zero-shot} settings.

\begin{figure} [t]
    \centering
    \includegraphics [width=0.85\linewidth]{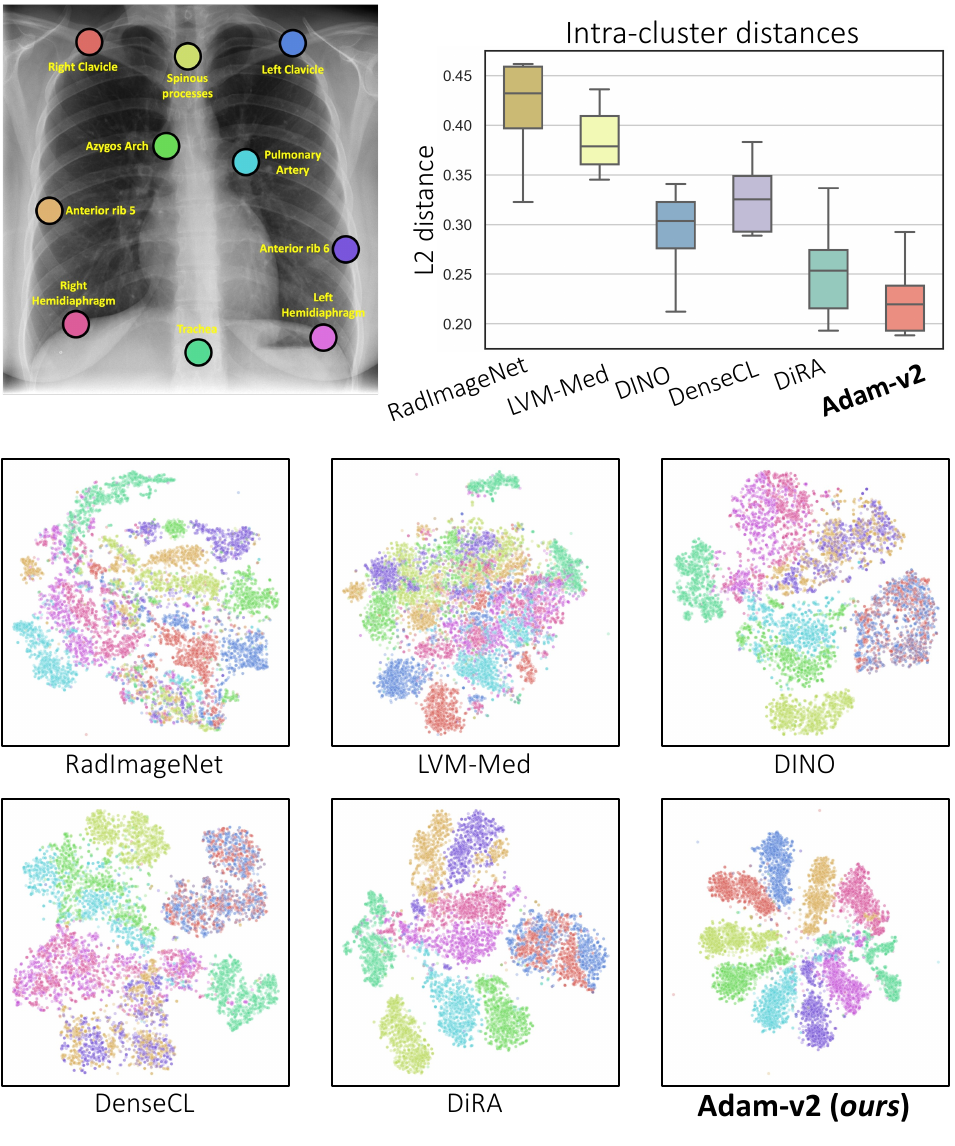}
    \caption{Adam--v2 learns localizability of anatomical structures, providing discriminative features for different landmarks. Same-colored points are instances of the same landmark across images.}
    \label{fig:locality}
\end{figure}

\begin{figure} [t]
    \centering
    \includegraphics [width=0.86\linewidth]{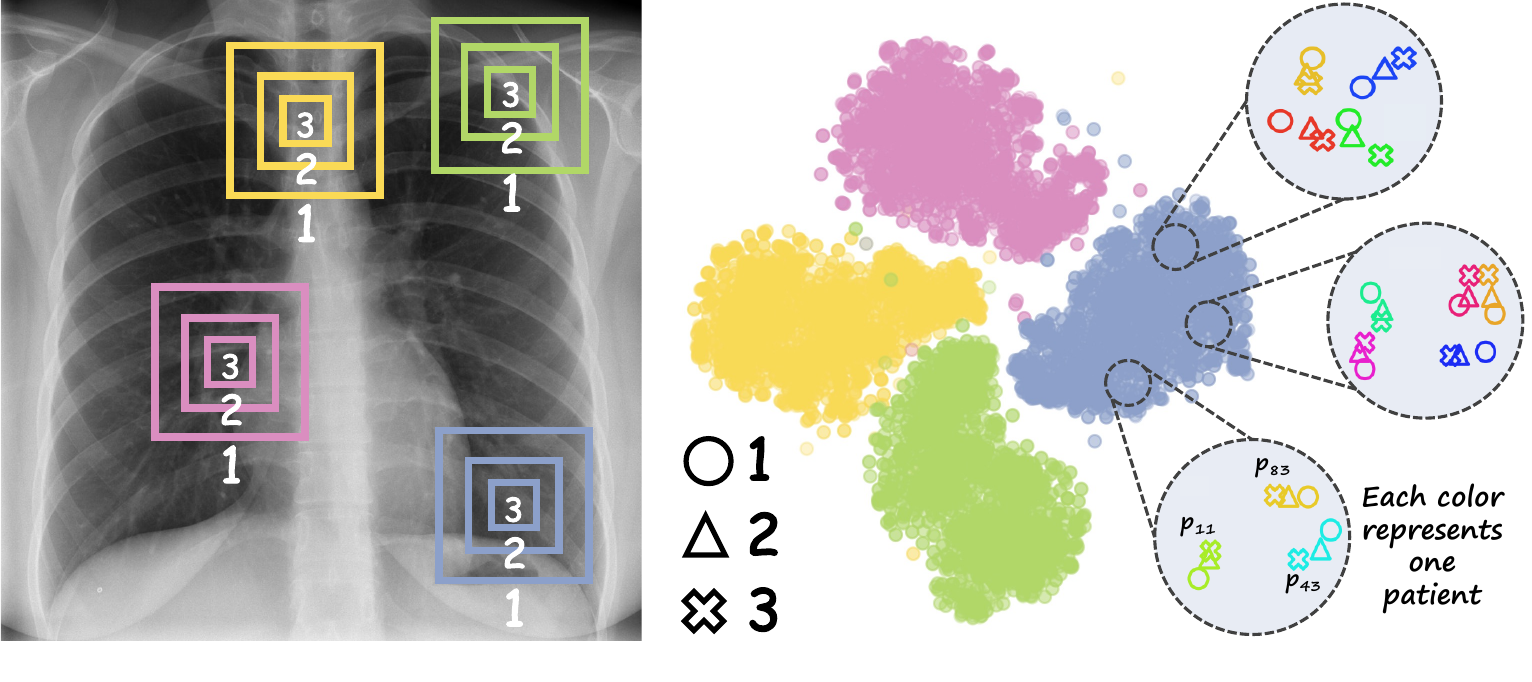}
    \caption{Adam--v2  balances diversity and harmony in embeddings of similar anatomical structures across patients and scales.\vspace{-4pt}
    }
    \label{fig:hierachy}
\end{figure}

\smallskip
\noindent\textbf{(1) \textit{Localazability:}} We investigate Adam--v2’s capability in discriminating different anatomical structures to determine if the learned embeddings (Eve--v2) preserve the locality of anatomical structures. To do so, we create a dataset of 1,000 images (from the ChestX-ray14 dataset) with 10 distinct anatomical landmarks manually annotated by human experts in each image (see \cref{fig:locality}). We extract patches of size $224^2$ around each landmark’s location across images and extract latent features of each landmark instance using each pretrained model under study (with \textit{no} fine-tuning). We then visualize the embeddings with t-SNE~\cite{Laurens2008Journal} plot. We compare
Adam--v2 with the RadImageNet, LVM-Med, and a representative set of SSL methods. As seen in \cref{fig:locality}, the baselines fall short in generating distinct features for different landmarks, leading to ambiguous embedding spaces with mixed clusters. By contrast, our Adam--v2 effectively discriminates between various anatomical landmarks, resulting in well-separated clusters within its learned embedding space. We complement our qualitative results (t-SNE plots) with quantitative results (box plots) by calculating intra-cluster distance for each landmark class and visualizing  the distances distributions with boxplots in \cref{fig:locality}. As seen, our Adam--v2 exhibits lower median distances, indicating more cohesive clusters, compared to the baselines. To showcase Adam--v2's capacity in balancing anatomical diversity and harmony and conveying hierarchical relationships, we randomly select four distinct anatomical landmarks, extract three patches of different resolutions (labeled as levels 1, 2, and 3) around each landmark across the images, and compute their embeddings with Adam--v2’s pretrained model. As seen in~\cref{fig:hierachy}, the embeddings of anatomical structures at levels 1, 2, and 3 for each landmark are closely aligned, highlighting Adam--v2's capability to preserve harmony in embeddings of semantically similar anatomical structures across resolutions and patients. Also, within each landmark, the embeddings of patches with levels 1, 2, and 3 for the same patient  (color-coded in~\cref{fig:hierachy}) are close, while those of different patients are well separated, representing Adam--v2's capability to preserve diversity of anatomical structures across patients.

\begin{figure} [t]
    \centering
    \includegraphics [width=0.88\linewidth]{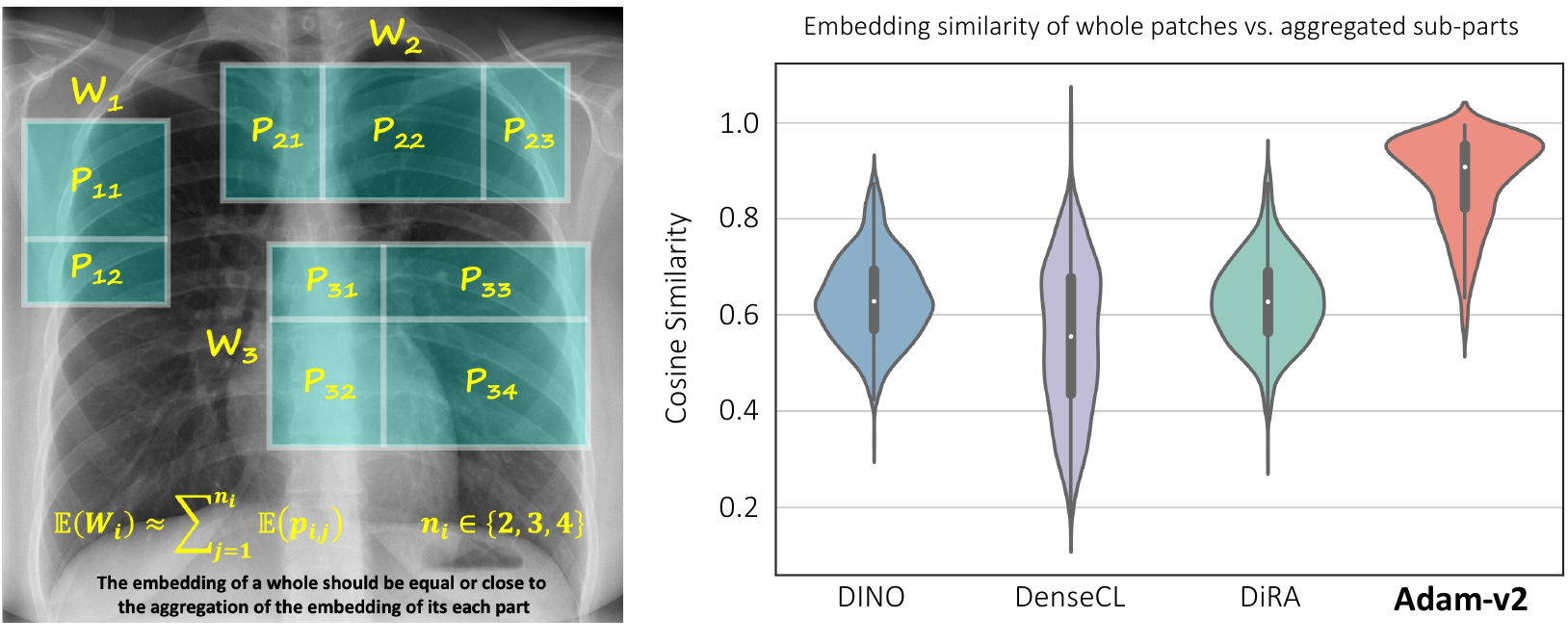}
    \caption{Adam--v2's embeddings (Eve--v2) encode part-whole relations of anatomical structures. \vspace{-4pt}}
    \label{fig:composition}
\end{figure}

\begin{figure} [t]
    \centering
    \includegraphics [width=0.88\linewidth]{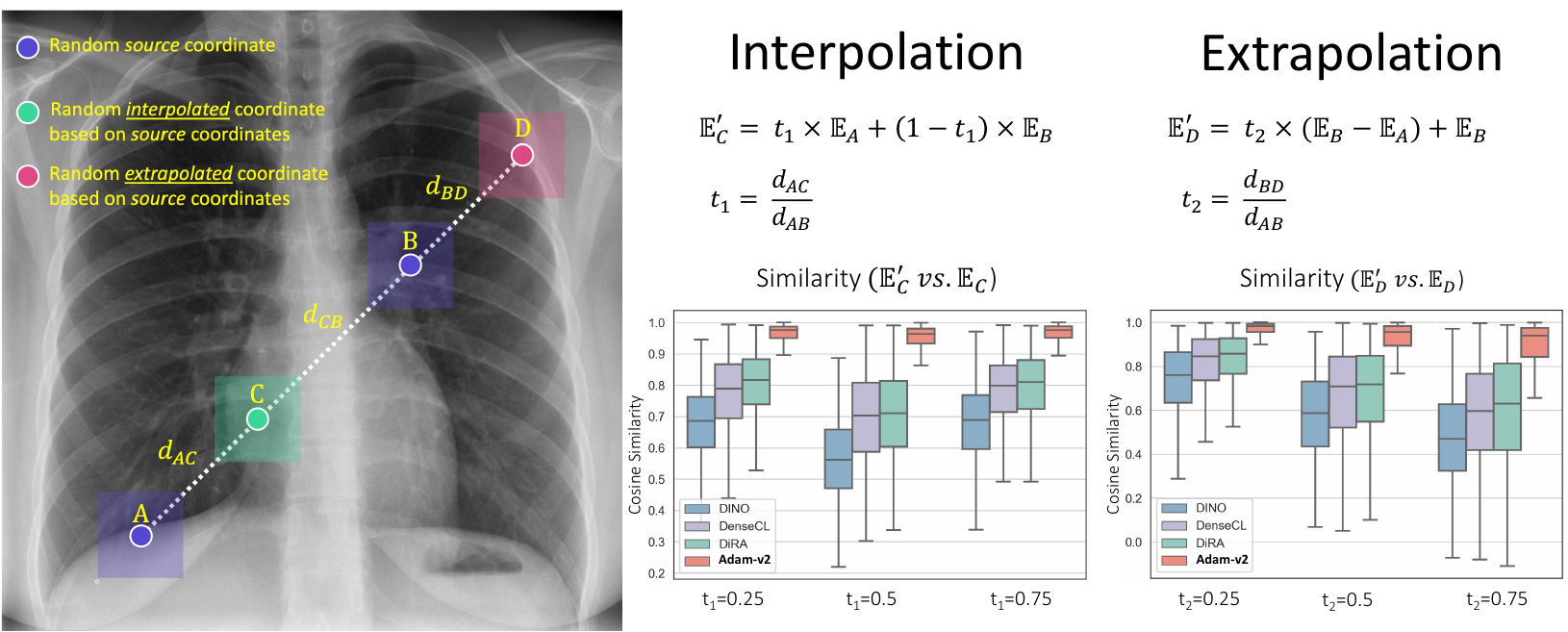}
    \caption{Adam--v2 exhibits two emergent properties: Interpolation \& Extrapolation. For interpolation/extrapolation, similarity has been computed between the interpolated/extrapolated embeddings ($E'_C/E'_D$) and their corresponding ground truth ($E_C/E_D$). \vspace{-3pt}}
    \label{fig:extra_intra}
\end{figure}

\begin{table*}[t]
 \centering
\footnotesize
\scalebox{0.85}{
\begin{tabular}{l|cccc|cccc|cc|cc}
\toprule
\multirow{3}{*}{Method} & \multicolumn{8}{c|}{Anatomical Structure Segmentation }& \multicolumn{4}{c}{Disease Segmentation} \\
 & \multicolumn{4}{c|}{JSRT-Clavicle (Dice\%)} & \multicolumn{4}{c|}{JSRT-Heart (Dice\%)} & \multicolumn{2}{c|}{SIIM-ACR (Dice\%)} & \multicolumn{2}{c}{ChestX-Det (IoU\%)} \\
\cline{2-5} \cline{6-9} \cline{10-11}  \cline{12-13}
 & 			3-shot &	6-shot&	12-shot  &  24-shot & 3-shot &	6-shot&	12-shot  &  24-shot & 5\% &	10\% &	5\% &	10\%    \\
\shline
RadImageNet~\cite{Mei2023RadImageNet} &  55.52&	71.26&	82.57&	83.29 & 73.12&	75.42&	89.22&	\underline{91.00}& \underline{54.56}&61.48& 64.22 &67.10\\
LVM-Med~\cite{nguyen2023lvmmed} & \underline{56.87}&	\underline{72.99}&	\underline{83.48}&	\underline{84.10}& \underline{79.45}&	\underline{86.94}&	\underline{89.98}&	90.78& 54.13& \underline{62.31} &\underline{65.11} & \underline{67.14} \\
\hline
DINO~\cite{Caron2021Emerging} &  24.06 & 29.59 & 38.54 & 45.01 &  45.45 & 60.79& 70.85& 80.78&47.85 &	52.08 &46.84	&52.64 \\
DenseCL~\cite{Wang2021Dense} & \underline{36.43}&	\underline{51.31}&	63.03&	69.13&\underline{64.88}&	\underline{74.43}&	75.79&	80.06&\underline{48.07}&	\underline{52.32}&	 60.18&	\underline{65.76}\\ 
DiRA~\cite{Haghighi2022DiRA} & 31.42&	38.59&	\underline{66.81}&	\underline{73.06}&63.76&	64.47&	\underline{76.10}&	\underline{81.42} &42.44& 48.27 	& \underline{61.63}&	64.86  \\
\textbf{Adam--v2 (Ours)} & \textbf{73.59}&	\textbf{79.57}&	\textbf{84.00}&	\textbf{85.96}& \textbf{86.88}&	\textbf{89.87}&	\textbf{90.47}&	\textbf{91.39} & \textbf{55.61}	&\textbf{68.11}	& \textbf{65.92}	&\textbf{68.17} \\
${\Delta_{\tiny 1}}$  & \textcolor{ForestGreen}{+16.7} & \textcolor{ForestGreen}{+6.58} & \textcolor{ForestGreen}{+0.52} & \textcolor{ForestGreen}{+1.86}  &\textcolor{ForestGreen}{+7.43} & \textcolor{ForestGreen}{+2.93} & \textcolor{ForestGreen}{+0.49} & \textcolor{ForestGreen}{+0.39} & \textcolor{ForestGreen}{+1.05} & \textcolor{ForestGreen}{+6.50} & \textcolor{ForestGreen}{+0.81} & \textcolor{ForestGreen}{+1.03}\\
${\Delta_{\tiny 2}}$  &\textcolor{ForestGreen}{+37.1} & \textcolor{ForestGreen}{+28.2} & \textcolor{ForestGreen}{+17.2} & \textcolor{ForestGreen}{+12.9} &  \textcolor{ForestGreen}{+22.0}&\textcolor{ForestGreen}{+15.4} & \textcolor{ForestGreen}{+14.3} & \textcolor{ForestGreen}{+9.97} & \textcolor{ForestGreen}{+7.54} & \textcolor{ForestGreen}{+15.7} & \textcolor{ForestGreen}{+4.29} & \textcolor{ForestGreen}{+2.41}\\

\hline
\end{tabular}
}
\caption{Adam--v2 excels in few-shot transfer, outperforming large-scale medical models (RadImageNet and LVM-Med) and SSL baselines across segmentation tasks. ${\Delta_1}$ and ${\Delta_2}$ show
Adam-v2’s performance boosts over second-best large-scale and SSL baselines, respectively.}
\label{tab:fewshot}
\end{table*}

\smallskip
\noindent\textbf{(2) \textit{Composability} \& \textit{Decomposability}:} We explore Adam--v2’s ability to capture part-whole hierarchies, as imposed by the composability and decomposability branches, in its learned embeddings (Eve--v2). To do so, we extract random patches of varying sizes, called \textit{whole}, from ChestX-ray14 test images. Each \textit{whole} is decomposed into 2, 3, or 4 non-overlapping parts with different sizes. We resize each \textit{whole} and its parts to $224^2$, extract features using pretrained models, and calculate the cosine similarity between the embedding of each \textit{whole} and the aggregate of its parts. As seen in \cref{fig:composition}, the box plot elements indicate that the median similarity for our Adam--v2 is significantly higher than that of other SSL baselines. Additionally, the distribution of our Adam--v2's similarity values is highly concentrated around the 1.5x interquartile, situated at the top of the box plot. This concentration suggests that, in most cases, the similarity value between the embedding of entire \textit{wholes} and their aggregated parts is closer to 1 in our Adam--v2 model.

\begin{figure*} [t]
    \centering
    \includegraphics [width=0.95\linewidth]{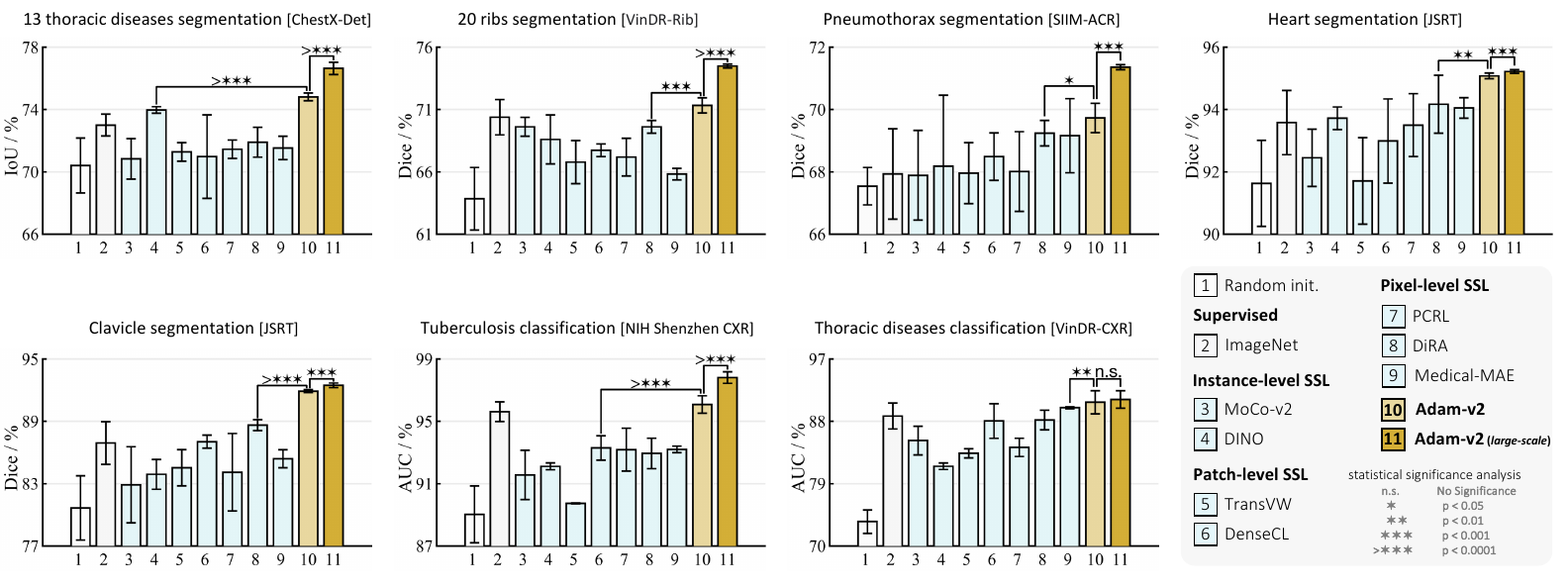}
    \caption{Adam--v2 provides generalizable and robust representations, outperforming SOTA self-supervised methods across diverse downstream tasks. Statistical significance analysis ($p < 0.05$) was conducted between Adam--v2 and the top SSL baseline in each task.}
    \label{fig:sota_full_finetune}
\end{figure*}

\smallskip
\noindent\textbf{(3) \textit{Interpolatation} }\textbf{\&} \textbf{(4) \textit{extrapolation}:} We investigate the Adam--v2's capability to interpolate/extrapolate embeddings for a randomly chosen anatomical structure by leveraging the embeddings of two other randomly selected anatomical structures. For interpolation, we select two random source coordinates (labeled as \(A\) and \(B\) in \cref{fig:extra_intra}) and use the established interpolation formula (refer to \cref{fig:extra_intra}) to interpolate a random point \(C\). We extract $224^2$ patches around points  A, B, and C and pass them through each pretrained model under study to extract their respective embeddings $E_A$, $E_B$, and $E_C$, where $E_C$ serves as the ground truth for evaluating  the interpolated embeddings for C.  Subsequently, we apply the  interpolation formula to generate embeddings for C based on $E_A$ and $E_B$, resulting in interpolated embeddings $E'_C$.  Finally, we compute the cosine similarity between the interpolated embeddings $E'_C$ and the ground truth $E_C$.  This process was repeated for 1,000 images selected from the test images of Chest X-ray 14, employing three different values of \(t_1\) (i.e., 0.25, 0.5, and 0.75). We use boxplots to illustrate the similarity distributions in each setting. We examine extrapolation of embeddings for a randomly selected point D in a similar manner using the extrapolation formula. 
The boxplots in  \cref{fig:extra_intra} reveal the consistent superiority of our Adam--v2  in delivering higher similarity between interpolated/extrapolated embeddings and the ground truth (with a median close to 1) compared to other baselines. This outstanding performance is indicative of the Adam--v2's capability in establishing relations between anatomical structures. It's noteworthy that our Adam--v2 model was \underline{\textit{not}} explicitly trained for these properties, and their emergence underscores the Adam--v2's capabilities in understanding anatomy.

\subsection{Adam--v2  excels in few-shot transfer, outperforming SOTA fully/self-supervised methods in segmentation tasks}
\label{sec:results_fewshot}

This section highlights the effectiveness of Adam--v2 as an effective foundation for fine-tuning deep models in segmentation tasks with limited labeled data. We compare Adam--v2 with 3 SSL methods, as well as RadImageNet and LVM-Med models, which serve as performance upper bounds. We conduct experiments on heart and clavicle segmentation tasks, fine-tuning the pretrained models using a few shots of labeled data (3, 6, 12, and 24) randomly sampled from JSRT dataset. Moreover, we conduct experiments on various thoracic disease segmentation tasks, fine-tuning the pretrained models on two randomly selected label fractions (5\% and 10\%) of the SIIM-ACR and ChestX-Det datasets. As seen in \cref{tab:fewshot}, our Adam--v2 outperforms both RadImageNet and LVM-Med across all label fractions in all tasks. For instance, in the 3-shot transfer for  clavicle and heart segmentation tasks, Adam--v2 surpasses LVM-Med by at least 16\% and 7\%, respectively. Moreover, Adam--v2 provides outstandingly better few-shot transfer performance compared with SSL methods across all tasks. For instance, in the pneumothorax segmentation task within the SIIM-ACR dataset, our Adam--v2 surpasses the runner-up baseline by 7.54\% and 15.7\% in the 5\% and 10\% labeled data subsets, respectively. Similarly, across the 5\% and 10\% fractions of the ChestX-Det dataset, our Adam--v2 demonstrates notably higher averages of 4.29\% and 2.41\% in the thoracic diseases segmentation task. Our attribution of Adam--v2’s superior representations for few-shot segmentation tasks is grounded in the significance of anatomy learning through our SSL approach and its profound impact on representation learning, which is neglected in existing methods.

\begin{table}[t]
\setlength{\tabcolsep}{7pt}
\begin{center}
\begin{threeparttable}[t]
\small
\begin{tabular}{lcc}

\toprule
Method &\# Pretraining Data & AUC\tnote{$\dagger$}   \\
\shline
RadImageNet~\cite{Mei2023RadImageNet} & 1.3M & 80.7\\
LVM-Med~\cite{nguyen2023lvmmed} & 1.3M & 82.0\\
Medical MAE~\cite{Xiao2023Delving}& 0.5M & 83.0\tnote{$\ddagger$}\\
\textbf{Adam--v2 (\textit{Large-scale})} & $\sim$1M & \textbf{83.4}\\

\hline
\end{tabular}
\begin{tablenotes}
\tiny
    \item[$\dagger$] We report mean AUC over 14 diseases on the official test split of ChestX-ray14 dataset.
    \item[$\ddagger$] We adopted this performance  reported by the original authors~\cite{Xiao2023Delving}; All the rest performance is ours.
    \end{tablenotes}
\end{threeparttable}
\caption{Adam--v2 outperforms previous SOTA methods (officially released large-scale medical vision models) on the public ChestX-ray14 benchmark, yielding a new record mAUC of 83.4\%. \vspace{-23pt}}
\label{tab:public_chestxray14}
\end{center}
\end{table}

\subsection{Adam--v2 stands out in full transfer, unleashing generalizable representations for a variety of tasks}
\label{sec:results_fullfinetune}
This section demonstrates the generalizability of Adam--v2’s representations via transfer learning to a broad range of downstream tasks in a full fine-tuning setting. We compare Adam--v2 with 7 SOTA ConvNet- and vision transformer-based SSL methods designed for both computer vision and medical applications. We include training downstream models from random initialization (the lower-bound baseline) and fully-supervised ImageNet model. As seen in \cref{fig:sota_full_finetune}, our Adam--v2 consistently achieves superior performance compared with the fully-supervised ImageNet model, as well as significant performance boosts ($p < 0.05$) compared with all SSL counterparts across all tasks.

\smallskip
\noindent\textit{\textbf{Comparison in Public ChestX-ray14 Benchmark.}} To scrutinize the scalability of our framework, we pretrained Adam--v2 with the ConvNeXt-B backbone on nearly 1M chest X-ray images and compared it against officially released large-scale medical vision models in the ChestX-ray14 benchmark. As seen in \cref{tab:public_chestxray14}, Adam--v2 hits a new record of 83.4 in the ChestX-ray14 benchmark. This suggests that a meticulously crafted learning strategy that comprehends human anatomy can fully harness large-scale data, thereby paving the way for developing powerful self-supervised models foundational to medical imaging.

\subsection{Ablation Experiments}
\noindent\textbf{\textit{Generalizability of our framework.}}  Our framework can seamlessly extend to other imaging modalities. To demonstrate this, we consider fundus images and pretrain Adam--v2 using the EyePACS dataset and then fine-tune it for two downstream tasks, considering both low-data regimes and full fine-tuning settings. As seen in \cref{tab:fundus},  Adam--v2 exhibits superior performance ($p < 0.05$) across tasks in both settings compared with SSL baselines that leverage the same pretraining data as our Adam--v2. Moreover, Adam--v2 outperforms ($p < 0.05$) RadImageNet and LVM-Med models in low-data regimes and achieves superior or equivalent performance in full fine-tuning scenarios.

\begin{figure} [t]
    \centering
    \includegraphics [width=1\linewidth]{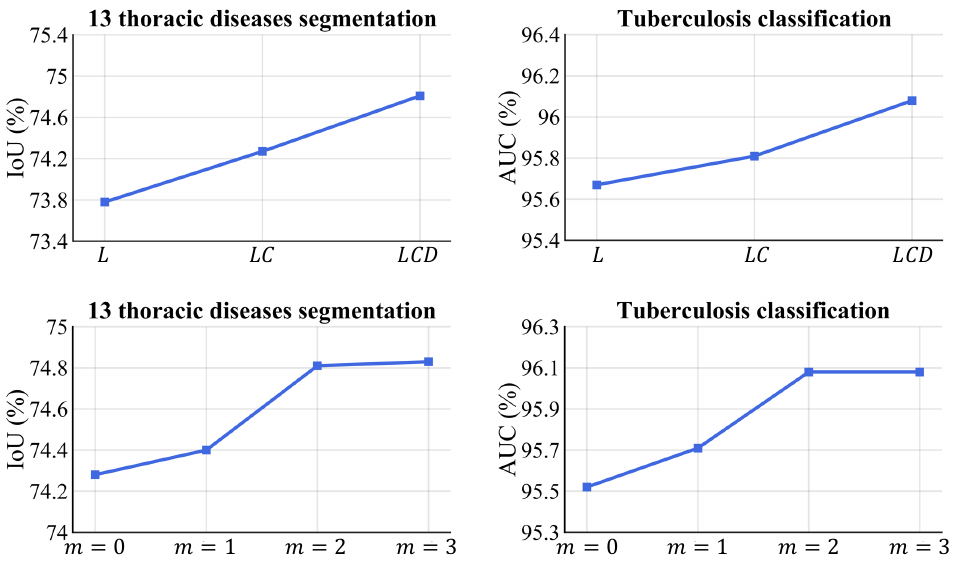}
    \caption{Ablation on the impact of (a) different branches of Adam--v2 (top-row) and (b) coarse-to-fine learning (bottom-row).}
    \label{fig:ablation_loss_gran}
\end{figure}

\noindent\textbf{\textit{Effect of learning objectives.}} We assess the impact of each learning branch in Adam--v2 by starting from localizability and incrementally
adding composability and decomposability learning. We fine-tune the models for two downstream tasks. As seen in the top-row of \cref{fig:ablation_loss_gran}, augmenting localizability with composability learning consistently improves performance across tasks. Moreover, the inclusion of decomposability further enhances the performance, resulting in significant performance boosts ($p < 0.05$)  in both tasks compared to standalone localizability learning.

\noindent\textbf{\textit{Effect of coarse-to-fine learning.}} We investigate the impact of hierarchical learning of anatomical structures at various scales (i.e. $m$) by initially training Adam--v2 with the entire anatomy ($m=0$) and then progressively delving deeper into the higher levels of anatomy hierarchy (up to level 3), representing finer anatomical structures. As seen in bottom-row of \cref{fig:ablation_loss_gran}, gradual increment of data granularity from $m=0$ to $m=2$ consistently improves the downstream performance. This highlights that our coarse-to-fine learning strategy incrementally deepens the model's anatomical knowledge, resulting in more generic representations for myriad tasks. Additionally, no significant change in performance is observed at $m=3$, suggesting that pretraining up to level 2 yields sufficiently robust representations. 

\begin{table}[t]
 \centering
\footnotesize
\scalebox{0.91}{
\begin{tabular}{lcc|cc}

\toprule
\multirow{2}{*}{Method} &  \multicolumn{2}{c|}{DRIVE (Dice\%)} & \multicolumn{2}{c}{Drishti-GS (Dice\%)} \\
\cline{2-3} \cline{4-5}
& 10\% & 100\% & 10\% & 100\% \\
\shline
Random & 74.03	\tiny{(0.87)}& 78.27 \tiny{(0.40)} & 70.17	\tiny{(10.91)}&94.53	\tiny{(1.72)} \\
\hline
RadImageNet & 76.53	\tiny{(0.49)} & 78.55	\tiny{(0.17)} & 90.37	\tiny{(1.48)} & 96.33	\tiny{(0.15)}\\
LVM-Med & 77.19	\tiny{(0.75)} &79.46	\tiny{(0.14)} &91.60	\tiny{(2.19)} & \textbf{97.02	\tiny{(0.15)}}\\
\hline
DINO & 75.89	\tiny{(0.63)} &78.36	\tiny{(0.28)} &85.90	\tiny{(3.27)} & 96.44	\tiny{(0.33)}\\
DenseCL &75.76	\tiny{(0.90)} &78.36	\tiny{(0.47)} &86.04	\tiny{(3.27)}& 96.60	\tiny{(0.01)}\\ 
DiRA &75.92	\tiny{(0.90)} &78.52	\tiny{(0.38)} &91.19	\tiny{(1.86)} & 96.76	\tiny{(0.16)}\\

Adam--v2 (Ours) & \textbf{78.04	\tiny{(0.14)}}$\blue{^{\pmb{\star}}}$&\textbf{79.91	\tiny{(0.18)}}$\blue{^{\pmb{\star}}}$ & \textbf{94.04	\tiny{(0.56)}}$\blue{^{\pmb{\star}}}$&\textbf{97.02	\tiny{(0.19)}}\\

\bottomrule
\end{tabular}
}
\caption{Adam--v2 outperforms SSL methods in fundus downstream tasks. $\blue{\pmb{\star}}$ shows statistically significant ($p < 0.05$) boosts.}
\label{tab:fundus}
\end{table}

%% file: sec/5_related_work.tex
\section{Related Work}
\label{sec:related}

\noindent\textbf{Self-supervised learning.}  A large body of SSL methods seek to learn global features via instance discrimination pretext tasks. These methods align the features of  augmented views from the same image by employing diverse learning objectives, including \textit{contrastive learning}~\cite{Ye2019Unsupervised,chen2020big,He2020Momentum,Chen2020Simple,chen2020improved,caron2021unsupervised,li2021prototypical,Chuang2022Robust,Guo2022HCSC,Zhang2022Dual,Chen2021Empirical},  \textit{self-distillation} ~\cite{Chen2021Exploring,Grill2020Bootstrap,Caron2021Emerging,Gidaris2021OBoW,Song2023Multi,Cao2023Guidelines,Caron2021Emerging}, and \textit{feature decorrelation} ~\cite{zbontar2021barlow,Bardes2021VICReg,Ermolov2021Whitening,bardes2022vicreg,zhang2022zerocl}. Alternatively, dense SSL methods seek to learn local features by encoding visual patterns embedded at smaller image regions. \textit{Dense contrastive learning} methods~\cite{Wang2022Exploring,Yang2021Instance,Zhang2022Leverage} enforce consistency between pixels at the same spatial location~\cite{Pinheiro2020Unsupervised,Xie2021Propagate,Bardes2022VICRegL}, similar pixels/patches in a feature map~\cite{Wang2021Dense,Bardes2022VICRegL}, or similar image regions~\cite{Xu2022RegionCL,Xiao2021Region,Xie2021DetCo,zhang2023patchlevel}. On the other hand, \textit{masked image modeling} methods~\cite{Xie2022SimMIM,He2022Masked,Chen2023Mixed,Feng2023Evolved,Bai2023Masked,Kong2023Understanding,Xie2023Data,Wang2023Hard,Liu2023MixMAE,Wang2023MaskedReconstruction,jiang2023layer,mishra2022simple,Tao2023Siamese} mask random portions of the images and  reconstruct the missing parts at pixel-level. Motivated by the success in computer vision, a broad variety of instance discrimination~\cite{Azizi2023Robust,azizi2021big, Kaku2021Intermediate} and image reconstruction methods~\cite{chen2019self,Xiao2023Delving}, along with their integration~\cite{Zhou2021Preservational,Taher2022CAiD,Haghighi2022DiRA,Tang2022Self}, have been explored for medical imaging. Given such advancements, the evolution of SSL has empowered it to serve as the cornerstone for developing foundation models with broad applicability~\cite{Bommasani2021Opportunities}. However, existing SSL methods overlook  anatomy hierarchies in their learning objectives, thereby lacking anatomy understanding capabilities. By contrast, Adam--v2 exploits the hierarchical nature of anatomy to learn semantics-rich features, leading to more pronounced models tailored for medical tasks.

\noindent\textbf{Learning from anatomy.} 
Consistent anatomy in medical imaging provides strong yet free supervision signals for deep models to learn common anatomical representations via self-supervision~\cite{zhou2019models}. Existing works revolve around recovering anatomical patterns from transformed images~\cite{zhou2019models,ZHOU2021Models}, learning semantics of recurrent anatomical patterns across patients~\cite{haghighi2020learning,haghighi2021transferable} with subsequent enhancements via adversarial learning~\cite{Haghighi2022DiRA,Guo2022Discriminative,Haghighi2024Self,Guo2024Discriminative}, exploiting  spatial relationships in anatomy~\cite{pang2022popar}, utilizing global and local anatomical consistency~\cite{zhou2023learning}, and incorporating anatomical cues to improve contrastive learning~\cite{Chaitanya2020Contrastive,Fu2022Anatomy,Hu2022Anatomy,jiang2023anatomical}. These existing works neglect hierarchical anatomy relations. Although our earlier method Adam~\cite{Taher2023Adam} uses anatomy hierarchies as soft supervisory signals, our Adam--v2 explicitly encodes part-whole hierarchies via its learning objectives. Compared with Adam~\cite{Taher2023Adam},  Adam--v2 showcases two significant advancements: (1) enhancing the localizability branch by eliminating negative pairs pruning, thereby improving computational efficiency for large-scale pretraining, (2) introducing two novel components: composability and decomposability, which are crucial for capturing part-whole hierarchies.

\noindent\textbf{Learning part-whole hierarchies.} Hierarchical representation learning is ingrained in architectures such as ConvNets~\cite{he2016deep,liu2022convnet} and hierarchical vision transformers (ViT)~\cite{Liu2021Swin}. But,  the multi-scale feature hierarchy of common neural networks does not explicitly align with the part-whole hierarchy in images, leading to the advent of new architectures for encoding part-whole hierarchies~\cite{Hinton2018matrix,Kosiorek2019Stacked}. Notably, GLOM~\cite{hinton2021represent} introduced a conceptual framework that utilizes attention to learn part-whole hierarchies, and subsequent works proposed ViT-based architectures to implement it~\cite{sun2022visual,Garau2022Interpretable}. By contrast, Adam--v2 goes beyond architecture design by introducing a 
new learning strategy that encodes the semantics of part-whole hierarchies into the embedding space through three explicit training objectives: localizability, composability, and decomposability.

%% file: sec/6_conclusion.tex
\section{Conclusion}
\label{conclusion}
We present a SSL framework Adam--v2 that enhances visual representations by creating a hierarchy of embeddings for different anatomical structures. The major novelty of our work is \textit{explicitly} enforcing part-whole hierarchies via three learning objectives. Our experiments highlight the effectiveness of Adam--v2 in various tasks, surpassing a range of baselines.  We also demonstrate the semantic richness of our learned representations, which stem from explicitly acquired or autonomously emerging unique properties. 

\section*{Acknowledgements}
 
 We acknowledge that Zhou et al.\ first hypothesized, observed, and illustrated the phenomena~\cite{zhou2023learning} that emerge from their PEAC's anatomical embedding space via co-segmentation~\cite{amir2021deep}, which were coined as ``echo-chambers'' earlier by~\citet{hinton2021represent}, but the ``echo-chambers'' shown in Fig.~1 produced by Mohammad Reza Hosseinzadeh Taher are what directly and automatically emerged from our Adam–v2’s embeddings (Eve--v2). This research has been supported in part by ASU and Mayo Clinic through a Seed Grant and an Innovation Grant, and in part by the NIH under Award Number R01HL128785. The content is solely the responsibility of the authors and does not necessarily represent the official views of the NIH. This work has utilized the GPUs provided in part by the ASU Research Computing and in part by the Bridges-2 at Pittsburgh Supercomputing Center through allocation BCS190015 and the Anvil at Purdue University through allocation MED220025 from the Advanced Cyberinfrastructure Coordination Ecosystem: Services \& Support (ACCESS) program, which is supported by National Science Foundation grants \#2138259, \#2138286, \#2138307, \#2137603, and \#2138296. The content of this paper is covered by patents pending.

%% file: sec/7_appendix.tex
\appendix
\section*{Supplementary Materials}

\section{Overview}
\begin{itemize}
    \item \textbf{Goal.} The primary goal of this paper is to highlight the importance of constructing a hierarchy of embeddings that can be autodidactically learned from anatomy and that can enhance the generalizability and robustness of representation learning.
    \item \textbf{Hypothesis.} We hypothesize that if deep neural networks can comprehend images akin to human perception---parsing them into part-whole hierarchies~\cite{Hinton1979SomeDO,HINTON1990Mapping,hinton2021represent}---their learned features would exhibit increased generalizability, robustness, and interpretability.
    \item \textbf{Challenge.} While deep neural networks excel in learning multi-level feature spaces, their limitation often lies in the absence of explicit coding for part-whole hierarchies, hindering a nuanced understanding of hierarchical relationships among objects and their constituent parts~\cite{hinton2021represent, mounir2023streamer}.
    \item \textbf{Solution.} We propose a novel self-supervised learning framework–called Adam-v2–that, \textit{without} requiring anatomy labeling, explicitly incorporates part-whole hierarchies into its learning objectives through three key branches---localizability, composability, and decomposability---in order to preserve a semantic balance of anatomical diversity and harmony in its learned embedding space (\S\textcolor{red}{2}).
     \item \textbf{Contributions.} In addition to higher generalizability and transferability of our Adam-v2's learned representations (Fig. \textcolor{red}{7} and Tab. \textcolor{red}{2}), our Adam-v2 proves to be an effective few-shot learner, making it a potent pretraining model for segmentation tasks with a scarcity of annotations (Tab. \textcolor{red}{1} and \ref{tab:fewshot_ssl_fundus}). Furthermore, we present a comprehensive set of quantitative and qualitative feature analyses that offer new perspectives for assessing anatomy understanding from various viewpoints (\S\textcolor{red}{4.1}).
     \item \textbf{Relation to GLOM.}  Hinton recently introduced the idea of ``GLOM''~\cite{hinton2021represent}, aiming to signify the importance of explicitly presenting part-whole hierarchies in a neural network. Inspired by the conceptual idea beneath GLOM, we propose Adam-v2 which is \textit{fundamentally} different from GLOM in several key aspects.  Firstly, GLOM is an \ul{\textit{imaginary system}} without practical experimentation,  whereas the Adam-v2 is a \ul{\textit{functioning system}} rigorously evaluated across 10 tasks in diverse settings. Secondly, GLOM is an idea for developing \ul{\textit{an ideal architecture}} for constructing hierarchical representations. However, our Adam-v2 offers \ul{\textit{a functional framework}} that takes a step towards achieving the overarching goal shared with GLOM---interpreting images as part-whole hierarchies akin to human vision systems---through a simple yet effective \ul{\textit{learning strategy}} that does not rely on labeled data. A recent line of research works has attempted to implement GLOM, such as~\cite{sun2022visual,Garau2022Interpretable}. However, our work diverges from this line of research in that Adam--v2 encodes the semantics of part-whole hierarchies into the embedding space through training with three explicit objectives: localizability, composability, and decomposability.
\end{itemize}

\begin{table}[t]
\setlength{\tabcolsep}{6pt}
\setlength{\extrarowheight}{1pt}
 \centering
\scalebox{0.92}{
\begin{tabular}{lcccc}

\toprule
\multirow{2}{*}{Method} & \multicolumn{4}{c}{DRIVE (Dice\%)} \\
\cline{2-5} 
 & 			3-shot &	6-shot&	12-shot & Full-shot     \\
\shline
DINO~\cite{Caron2021Emerging} &  \underline{68.96}	& \underline{71.49}&	72.73& 78.36 \\
DenseCL~\cite{Wang2021Dense} & 67.99	&70.65&	72.83& 78.36\\ 
DiRA~\cite{Haghighi2022DiRA} & 68.13	&70.80&	 \underline{72.88}&\underline{78.52}\\

\textbf{Adam--v2 (Ours)} &\textbf{74.87}&	\textbf{75.74}&	\textbf{76.27}&\textbf{79.91} \\
${\Delta}$  & \textcolor{ForestGreen}{+5.92} & \textcolor{ForestGreen}{+4.25} & \textcolor{ForestGreen}{+3.39} & \textcolor{ForestGreen}{+1.39} \\

\hline
\end{tabular}
}
\vspace{0.3 em}\\
\scalebox{0.92}{
\begin{tabular}{lcccc}

\hline
\multirow{2}{*}{Method} & \multicolumn{4}{c}{Drishti-GS (Dice\%)} \\
\cline{2-5} 
 & 			3-shot &	6-shot&	12-shot & Full-shot   \\
\shline
DINO~\cite{Caron2021Emerging}& 85.29&	92.36&	93.95& 96.44 \\

DenseCL~\cite{Wang2021Dense} &85.42&	92.51&	94.10& 96.60 \\ 
DiRA~\cite{Haghighi2022DiRA} &  \underline{86.42}&	 \underline{92.81}&	 \underline{94.14} & \underline{96.76} \\

\textbf{Adam--v2 (Ours)} &\textbf{ 94.00}&	\textbf{94.97}&	\textbf{96.20} & \textbf{97.02}\\
${\Delta}$  & \textcolor{ForestGreen}{+7.57} & \textcolor{ForestGreen}{+2.16} & \textcolor{ForestGreen}{+2.07} & \textcolor{ForestGreen}{+0.26} \\

\bottomrule
\end{tabular}
}
\caption{Adam--v2 demonstrates superior performance in few-shot transfer within the fundus modality, surpassing SOTA SSL methods by a large margin in the retinal vessel and optic disk segmentation tasks on the DRIVE and Drishti-GS datasets, respectively. Remarkably, with only 3 shots, Adam--v2 achieves 94\% and 97\% of its full training data performance in retinal vessel and optic disk segmentation tasks, respectively. ${\Delta}$ shows Adam--v2’s performance boosts compared with second-best method (underlined). All methods are pretrained on Eye-PACS dataset.}
\label{tab:fewshot_ssl_fundus}
\end{table}

\begin{figure*}[t]
    \centering
    \includegraphics [width=1\linewidth]{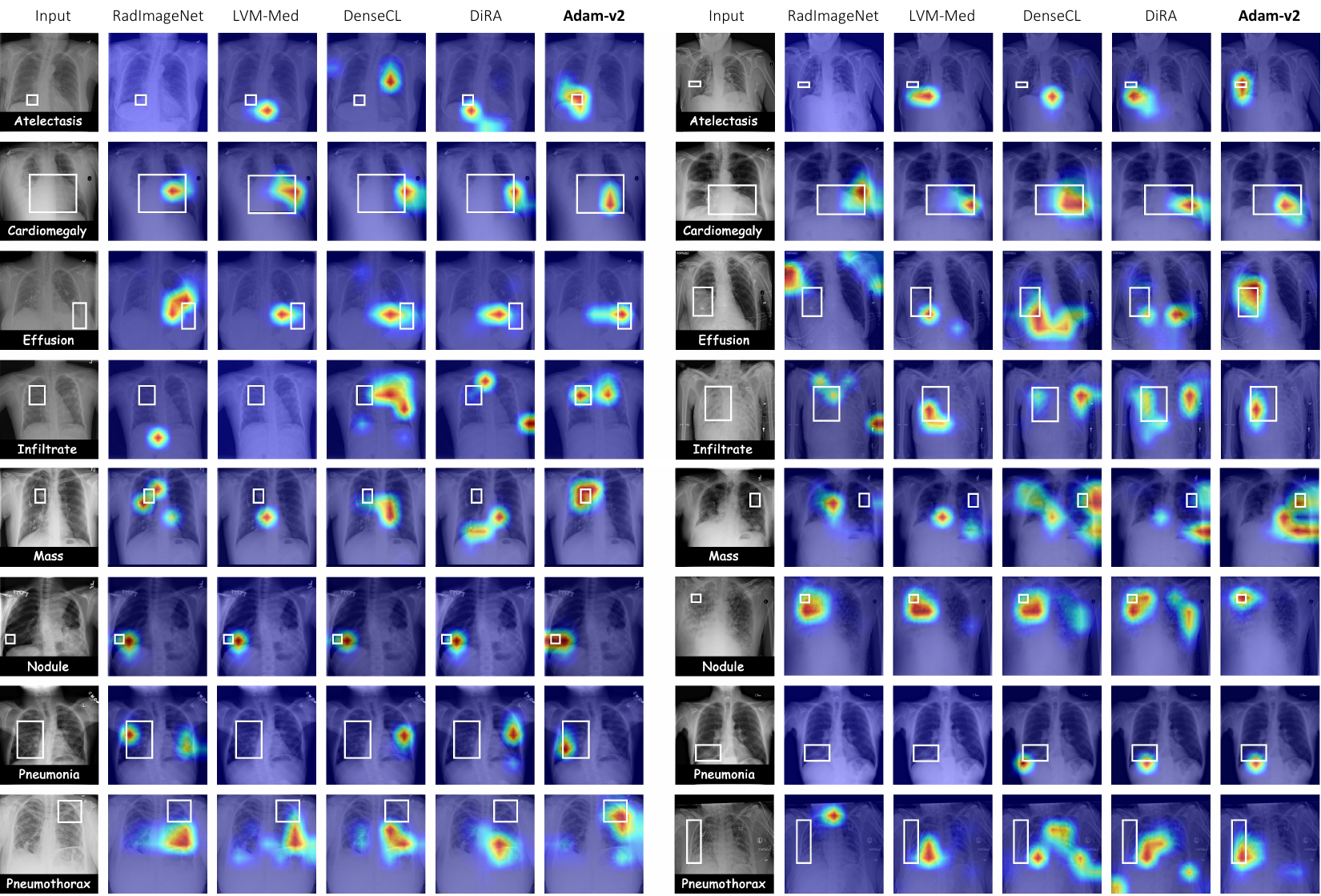}
    \caption{[Better viewed on-line, in color, and zoomed in for details] Visualization of Grad-CAM heatmaps generated by Adam--v2 and baseline methods for eight diseases in ChestX-ray14. Ground truth is marked with white boxes. Adam--v2 yields finer localization outcomes compared to baselines, which either concentrate on broader image regions or miss alignment with the ground truth.}
    \label{fig:weakly}
\end{figure*}

\section{Additional Results}
\subsection{Few-shot Transfer in Fundus Imaging Tasks}
To underscore the effectiveness of our SSL framework in learning robust representations for few-shot segmentation tasks, we replicate the experiments reported in Tab. \textcolor{red}{1} within few-shot transfer settings for fundus applications. To do so, we fine-tune our Adam--v2 model and baseline models, all of which were pretrained on fundus images from the EyePACS dataset, using a limited number of labeled samples (3-shot, 6-shot, and 12-shot) from the DRIVE and Drishti-GS datasets. As seen in \cref{tab:fewshot_ssl_fundus}, Adam--v2 exhibits superior few-shot transfer performance when compared to SSL methods in retinal vessel segmentation and optic disk segmentation tasks on the DRIVE and Drishti-GS datasets, respectively. Notably, when compared to the runner-up baseline, Adam--v2 achieves improvements of 5.92\%, 4.25\%, and 3.39\% in 3-shot, 6-shot, and 12-shot scenarios for the retinal vessel segmentation task, and 7.57\%, 2.16\%, and 2.07\% for the optic disk segmentation task. With only 3 shots, Adam--v2 achieves 94\% and 97\% of its full training data performance in retinal vessel and optic disk segmentation tasks, respectively.

\subsection{Weakly-supervised Disease Localization}

We explore the effectiveness of our Adam--v2 in localizing chest pathology in a weakly supervised setting. To do so, we follow~\cite{Xiao2023Delving, wang2017chestx} and leverage the ChestX-ray14 dataset, which comprises 787 cases annotated with bounding boxes for eight thorax diseases: Atelectasis, Cardiomegaly, Effusion, Infiltrate, Mass, Nodule, Pneumonia, and Pneumothorax. During the training phase, we initialize target models with our Adam--v2 and other baselines pretrained weights and fine-tune them using only image-level disease labels. In the testing phase, we employ Grad-CAM~\cite{jacobgilpytorchcam} to visualize image regions responsible for the model predictions, specifically identifying the diseased regions. As seen in~\cref{fig:weakly}, Adam--v2 localizes diseases more accurately compared to other baselines. Notably, the heatmaps generated by our Adam--v2 exhibit a higher degree of concentration around disease regions in comparison to other baselines, with its attention maps displaying a more pronounced overlap with the ground truth across all diseases. This generation of more interpretable activation maps not only highlights the Adam--v2's potential for precise disease localization but also shows its potential for clinical utility in post-hoc interpretation by radiologists.

\begin{figure*}[t]
    \centering
    \includegraphics [width=1\linewidth]{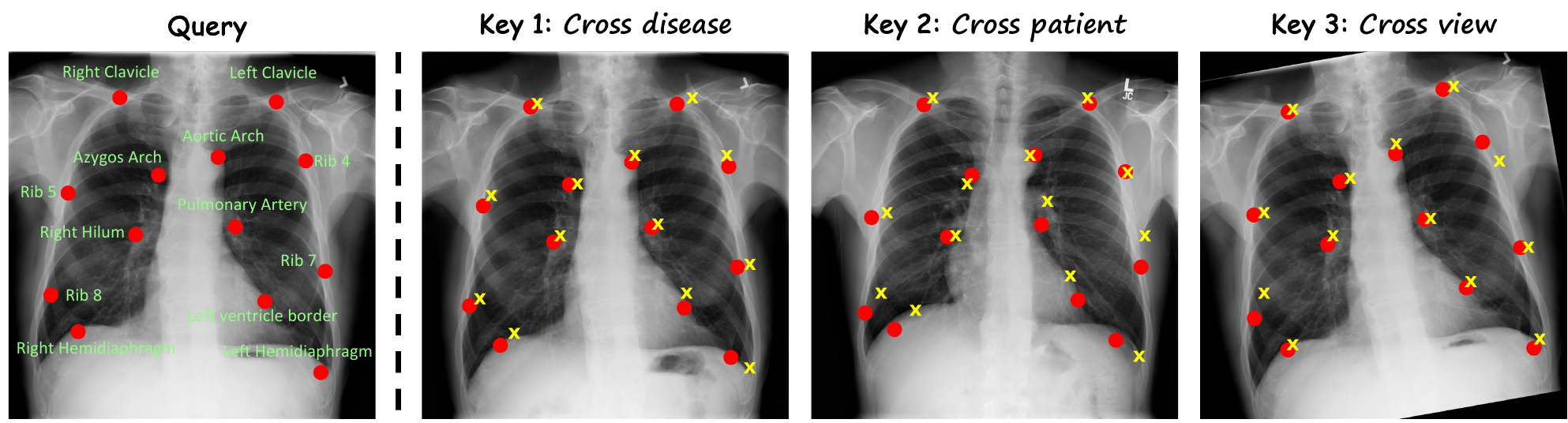}
    \caption{[Better viewed on-line, in color, and zoomed in for details] Adam--v2 shows its potential in identifying similar anatomical landmarks across three distinct settings: anatomical point matching across images of the same patient with different diseases, images of different patients, and augmented views of the same image. The images are from the test set of the ChestX-ray14 dataset. The red circles represent the ground truth for 13 distinct landmark points in query and key images, while the yellow crosses indicate the corresponding matched points identified by our Adam--v2.}
    \label{fig:landmark_detection}
\end{figure*}

\subsection{Anatomy Matching}
To further illustrate Adam--v2's capability in anatomy understanding, we examine Adam--v2's representations for anatomical landmark matching in a zero-shot setting. To do so, from a given query image, we randomly select $N_q=13$ anatomical landmark points and extract a patch of size $96^2$ centered at each anatomical landmark point. These patches are resized to $224^2$ and passed through Adam--v2's pretrained backbone to generate query embeddings $E_q=\{E_q^i\}_{i=1}^{N_q}$. Then, for an \textit{unlabeled} key image, we extract $N_k$ patches by sliding a window of size $96^2$ with a stride of 16. After resizing these key patches to $224^2$, their embeddings $E_k=\{E_k^j\}_{j=1}^{N_k}$ are obtained using Adam--v2's pretrained backbone. Finally, for each query anatomical landmark embedding in $E_q$, we compute its $\ell_2$-distance with all embeddings in $E_k$ and identify the center of the patch corresponding to the embeddings with the minimum distance as the matched point for the anatomical landmark.

To showcase the robustness of Adam--v2's representations to anatomical variations, we explore three distinct settings involving anatomical point matching across  images of the same patient with different diseases, images of different patients, and augmented views of the same image. \cref{fig:landmark_detection} depicts the query image annotated with 13 landmark points (red circles) and corresponding matched points (indicated by yellow crosses) across different diseases, patients, and views. To assess the accuracy of matched points, we include the ground truth landmark points provided by human experts for key images (depicted as red circles). As shown in Figure \ref{fig:landmark_detection}, Adam--v2 showcases its potential in precisely identifying similar anatomical landmarks, aligning with our findings in Sec. \textcolor{red}{4.1} and emphasizing the semantic richness inherent in Adam--v2's representations. These findings underscore an additional emergent property of our Adam--v2, revealing its potential in identifying corresponding landmarks. It is crucial to emphasize that our primary focus in this study is to acquire generalizable and semantically rich representations through our proposed SSL learning strategy. For an in-depth exploration of Adam--v2's potential in landmark detection and image registration—topics beyond the scope of this paper—a more detailed investigation is warranted, which we defer to future work.

\section{Downstream Tasks}
We extensively evaluate the transferability of Adam--v2 pretrained models across a broad spectrum of 10 downstream tasks on nine publicly available datasets of chest X-ray and fundus modalities. These tasks assess the generalizability Adam--v2's representations  across  
various applications (binary classification, multi-label classification, organ/lesion segmentation), diseases (tuberculosis, pneumothorax, lung nodules, etc), anatomical structures (heart, clavicle, ribs, vessels, optic disk), and modalities (chest radiography and fundus photography). In the following, we elaborate on the specifics of downstream tasks incorporated in this paper.

\medskip
\noindent\textbf{Task 1: Clavicle segmentation.} This task entails pixel-level segmentation of the left and right clavicles. We use the Japanese Society of Radiological Technology (JSRT) dataset~\cite{vanginneken2006Segmentation,Shiraishi2000jsrt}, comprising 247 posterior-anterior chest radiographs with associated segmentation masks for clavicles. The dataset was divided into two folds, containing 124 and 123 images, respectively. We adhere to the official patient-wise data split, utilizing fold-1 for training  and fold-2 for testing. We use mean Dice score as the evaluation metric for assessing clavicle segmentation performance.

\medskip
\noindent\textbf{Task 2: Heart segmentation.} This task encompasses pixel-level segmentation of the heart. We use JSRT dataset~\cite{vanginneken2006Segmentation,Shiraishi2000jsrt} for this task, comprising 247 images with associated segmentation masks for heart. We adhere to the official patient-wise data split, utilizing fold-1 (124 images) for training  and fold-2 (123 images) for testing. We use mean Dice score as the evaluation metric for assessing heart segmentation performance.

\medskip
\noindent\textbf{Task 3: Ribs segmentation.} This task entails pixel-level segmentation of individual ribs. We utilize the VinDr-Rib dataset~\cite{Nguyen2021VinDr}, comprising 245 chest radiographs accompanied by segmentation masks for 20 individual anterior and posterior ribs (10 on each side of the lungs). Following the official dataset split, we use 196 images for training and 49 for testing. This task is formulated as a multi-class segmentation problem, and the performance is evaluated using the mean Dice score.

\medskip
\noindent\textbf{Task 4: Thoracic diseases segmentation.} This task involves pixel-level segmentation of thoracic diseases using ChestX-Det dataset~\cite{Lian2021Structure}. The dataset comprises 3,578 chest X-ray images. Board-certified radiologists have provided segmentation masks for 13 common thoracic conditions, including atelectasis, calcification, cardiomegaly, consolidation, diffuse nodule, effusion, emphysema, fibrosis, fracture, mass, nodule, pleural thickening, and pneumothorax. We adhere to the official dataset split, using 3,025 images for training and 553 images for testing.

\medskip
\noindent\textbf{Task 5: Pneumothorax segmentation.}   This task focuses on pixel-level segmentation of pneumothorax disease. We utilize the SIIM-ACR dataset~\cite{PNEchallenge}, consisting of 10,000 chest radiographs along with segmentation masks for pneumothorax disease, if present in an image. For training and testing, we randomly divide the dataset into 8,000 and 2,000 images, respectively. Segmentation performance is assessed using the mean Dice score.

\medskip
\noindent\textbf{Task 6: Common thoracic diseases classification.} This task involves multi-label classification of five common thoracic diseases. We use VinDR-CXR dataset~\cite{nguyen2020vindrcxr} that provides 18,000 posterior-anterior chest radiographs, along with image-level labels provided by expert radiologists for 6 conditions: lung tumor, pneumonia, tuberculosis, other diseases,  COPD, and No finding. Following the official dataset split, we allocate 15,000 images for training and 3,000 for testing, and evaluate classification performance using the mean AUC score. 

\medskip
\noindent\textbf{Task 7: Tuberculosis classification.} This task involves detection of tuberculosis disease. We use NIH Shenzhen CXR dataset~\cite{Jaeger2014Tow}, including 662 chest radiographs, with 326 images categorized as normal and 336 images representing patients with tuberculosis. We randomly split the dataset into training (80\%) and testing (20\%) sets, and evaluate performance using the AUC metric.

\medskip
\noindent\textbf{Task 8: Thoracic diseases classification.} This task encompasses multi-label classification of fourteen thoracic diseases, employing the ChestX-ray14 dataset~\cite{wang2017chestx} curated by the National Institutes of Health Clinical Center, USA. The dataset comprises 112,120 de-identified X-rays from 30,805 unique patients, with labels indicating the absence or presence of 14 thoracic disease categories. We adhere to the official patient-wise split provided by the dataset, allocating 86K images for training and 25K for testing, and assess classification performance using the mean AUC over the 14 diseases.

\medskip
\noindent\textbf{Task 9: Retinal vessel segmentation.} This task encompasses pixel-level segmentation of retinal vessels. We use DRIVE dataset~\cite{Budai2013Robust}, including 40 color fundus images along with expert annotations for retinal vessels. Following the official dataset split, we use 20 images for training and 20 for testing, and evaluate segmentation performance using the mean Dice score.

\medskip
\noindent\textbf{Task 10: Optic disk segmentation.} This task involves pixel-level segmentation of optic disk. We use Drishti-GS dataset~\cite{Sivaswamy2014Drishti}, encompassing 101 fundus images, divided into 50 training and 51 testing images. Ground truth segmentation masks are provided for optic disk by human experts. We adhere to the official dataset split and assess segmentation performance using the mean Dice score.

\section{Implementation Details}
\subsection{Pretraining Setup}
Our SSL framework is architecture-neutral and compatible with any ConvNet and vision transformer backbones. We have trained two Adam--v2 models with ResNet-50 backbone using unlabeled images from the training sets of ChestX-ray14~\cite{wang2017chestx} and EyePACS~\cite{Cuadros2009EyePACS} datasets for chest X-ray and fundus imaging tasks, respectively. Moreover, we have trained Adam--v2 with ViT-S backbone using unlabeled images from the training sets of ChestX-ray14 and CheXpert~\cite{irvin2019chexpert} datasets. Additionally, to demonstrate the scalability of our framework, we have trained a large-scale Adam--v2 model with ConvNeXt-B backbone using a large corpus of 926K chest X-ray images collected from 13 publicly-available datasets, including ChestX-ray14, CheXpert~\cite{irvin2019chexpert}, VinDr-CXR~\cite{nguyen2020vindrcxr}, NIH Shenzhen CXR~\cite{Jaeger2014Tow}, RSNA Pneumonia Detection Challenge~\cite{RSNA2018}, MIMIC-CXR~\cite{Johnson2019MIMIC}, PadChest~\cite{Bustos2020PadChest}, COVID-19 Radiography Database~\cite{Chowdhury2020Can}, Indiana ChestX-ray~\cite{IndianaDataset}, Mendeley-V2~\cite{mendelydataset}, COVIDx~\cite{covidixdataset}, JSRT~\cite{vanginneken2006Segmentation}, and NIH Montgomery~\cite{Jaeger2014Tow}.
Following~\cite{Caron2021Emerging}, the localizability heads $h_{\theta_{LS}}$ and  $h_{\theta_{LT}}$    consist of a 3-layer multi-layer perceptron (MLP) with hidden dimension 2048 and output dimension $K=65536$. The composability ($h_{\theta_{C}}$) and  decomposability ($h_{\theta_{D}}$) heads are 2-layer MLP with hidden dimension 2048. We use AdamW optimizer, and follow \cite{Caron2021Emerging} in learning rate scheduler and weight decay settings.
To empower the model with hierarchical anatomy learning, we train Adam--v2 in a coarse-to-fine manner, incorporating diverse anatomical structures at various scales. Starting with $m=0$, where the model is trained on the entire anatomy (whole images), we progressively reduce the scale of anatomical structures by factors based on powers of 2. Specifically, for input images with spatial resolution ${\scriptstyle(H\times W)}$, we randomly sample anatomical structures with resolutions of $(\frac{H}{2^m}\times\frac{W}{2^m})$, where $m\in\{1,2,...\}$, and utilize them as inputs to the model during the pretraining process.  For learning anatomical structures at each scale, the model is trained with the objective function in Eq. (\textcolor{red}{5}). In practice, we assess anatomical structure resolutions across up to 4 scales (i.e., $m\in\{0,..,3\}$), but our ablation study (see Fig. \textcolor{red}{8}) suggests that up to three levels are sufficient to yield robust representations. During the training, the parameters of the teacher network and localizability head $h_{\theta_{LT}}$ are updated with an exponential moving average on the weights of the student network and $h_{\theta_{LS}}$, respectively; the update rules are $\theta_T \leftarrow \lambda\theta_T + (1-\lambda) \theta_S$ and $\theta_{LT} \leftarrow \lambda\theta_{LT} + (1-\lambda) \theta_{LS}$, where $\lambda$ follows a cosine schedule from 0.996 to 1 during training~\cite{Grill2020Bootstrap}. Following \cite{Caron2021Emerging}, we use centering and sharpening for the teacher's outputs to avoid collapsing solutions for localizability learning. In the localizability branch, we extract one global crop of size $224^2$ from the input $w$, along with eight multi-scale crops of size $96^2$. The temperature $\tau_s$ is set to 0.1, and $\tau_t$ follows a linear warm-up as \cite{Caron2021Emerging}. 
We train ResNet-50 model from scratch with a batch size 512 distributed across 8 Nvidia V100-32Gb GPUs. We first warm up the localizability branch with a scheme as \cite{Caron2021Emerging}  (200/200/100 epochs for $m=0,1,2$), empowering the model with an initial ability  to discriminate different anatomical structures. Subsequently, the composability and decomposability losses are integrated into the training process, and the entire framework is jointly trained  (10/90/165 epochs for $m=0,1,2$). We initialize the ConvNeXt-B model with ImageNet-22K pretrained weights, and train it with a batch size 160. We first warm up the localizability branch (70/70/30 epochs for $m=0,1,2$), and then train the model with all three losses (10/50/30 epochs for $m=0,1,2$). 

\subsection{Downstream Setup}
\noindent\textbf{Evaluations.} We utilize the pretrained teacher backbone of Adam--v2 (i.e., $g_{\theta_T}$) for zero-shot, few-shot, and full transfer evaluations. We use Adam--v2 with ResNet-50 backbone in zero-shot, few-shot, and full transfer evaluations. We use Adam--v2 model with ConvNeXt backbone for comparison with large-scale medical models in public ChestX-ray14 benchmark. We use Adam--v2 with ViT-S backbone for co-segment visualizations (Fig. \textcolor{red}{1}), and follow the settings of \cite{amir2021deep}  to co-segment the common chest anatomical structures. 

\noindent\textbf{Fine-tuning settings.} For transfer learning to segmentation tasks, we employ a U-Net architecture~\cite{Ronneberger2015Unet}, initializing the encoder weights with Adam--v2's pretrained backbone. For transfer learning to classification tasks, we take Adam--v2's pretrained backbone and append a fully connected layer to generate the desired classification outputs. Following the standard evaluation protocol~\cite{Taher2021Systematic,Taher2023Adam}, we perform end-to-end fine-tuning for all parameters of the target models across all downstream tasks. We strive to optimize each downstream task with the most effective hyperparameters.  In classification tasks, we use AdamW optimizer with learning rate $2.5e-4$ decayed by a cosine schedule, weight decay 0.05, $(\beta_1,\beta_2)=(0.9, 0.95)$, and standard data augmentation, encompassing random crop, flip, and rotation. In segmentation tasks, we use Adam optimizer with learning rate $1e-3$ for VinDR-Ribs, DRIVE, and Drishti-GS datasets, and AdamW optimizer with a learning rate $2e-4$ for the rest of the tasks, cosine learning rate decay scheduler, and standard data augmentation, encompassing random crop, brightness contrast, grid and optical distortion, elastic transformation, gamma.  Moreover, we employ early-stopping using 10\% of the training data as the validation set. We use input size $512^2$ for DRIVE and $224^2$ for all other tasks. We follow~\cite{Taher2021Systematic} for ChestX-ray14 dataset. Classification and segmentation performances are measured by the AUC (area under the ROC curve), and mean Dice coefficient metrics and IoU (Intersection over Union) metrics, respectively.

